%% file: main.tex
\documentclass[10pt,twocolumn,letterpaper]{article}
\usepackage[letterpaper,margin=0.72in,columnsep=0.28in]{geometry}

\usepackage{iftex}
\ifPDFTeX
  \usepackage[utf8]{inputenc}
  \usepackage[T1]{fontenc}
  \usepackage{tgpagella}
  \usepackage{textcomp}
  \usepackage{amssymb}          % \mathbb, which unicode-math supplies otherwise
\else
  \usepackage{fontspec}
  \usepackage{unicode-math}
  \IfFontExistsTF{texgyrepagella-math.otf}
    {\setmathfont{TeX Gyre Pagella Math}}
    {\setmathfont{Latin Modern Math}}
\fi
\usepackage{newunicodechar}

\usepackage{amsmath}
\usepackage{booktabs}
\usepackage{float}
\usepackage{graphicx}
\usepackage{adjustbox}
\usepackage{rotating}
\newlength{\rotmaxw}
\newlength{\rotmaxh}
\usepackage[font=small,labelfont=bf,skip=6pt]{caption}
\usepackage{xcolor}
\usepackage[verbose=silent]{microtype}
\usepackage{titlesec}
\usepackage{fancyhdr}
\usepackage{pifont}

\definecolor{accent}{RGB}{38,78,120}
\definecolor{linkblue}{RGB}{20,60,110}
\definecolor{urlblue}{RGB}{0,84,166}
\definecolor{headtint}{RGB}{24,42,64}

\usepackage[colorlinks=true,
            linkcolor=linkblue,
            citecolor=linkblue,
            urlcolor=urlblue,
            filecolor=urlblue,
            breaklinks=true]{hyperref}
\usepackage{xurl}

\ifPDFTeX
  \newcommand{\umap}[3]{\DeclareUnicodeCharacter{#2}{#3}}
\else
  \newcommand{\umap}[3]{\newunicodechar{#1}{#3}}
\fi
\umap{∈}{2208}{\ensuremath{\in}}
\umap{∉}{2209}{\ensuremath{\notin}}
\umap{∖}{2216}{\ensuremath{\setminus}}
\umap{∪}{222A}{\ensuremath{\cup}}
\umap{∩}{2229}{\ensuremath{\cap}}
\umap{⊆}{2286}{\ensuremath{\subseteq}}
\umap{⊂}{2282}{\ensuremath{\subset}}
\umap{ℝ}{211D}{\ensuremath{\mathbb{R}}}
\umap{∅}{2205}{\ensuremath{\emptyset}}
\umap{×}{00D7}{\ensuremath{\times}}
\umap{≤}{2264}{\ensuremath{\leq}}
\umap{≥}{2265}{\ensuremath{\geq}}
\umap{≈}{2248}{\ensuremath{\approx}}
\umap{≠}{2260}{\ensuremath{\neq}}
\umap{→}{2192}{\ensuremath{\rightarrow}}
\umap{↔}{2194}{\ensuremath{\leftrightarrow}}
\umap{−}{2212}{\ensuremath{-}}
\umap{±}{00B1}{\ensuremath{\pm}}
\umap{ρ}{03C1}{\ensuremath{\rho}}
\umap{τ}{03C4}{\ensuremath{\tau}}
\umap{Δ}{0394}{\ensuremath{\Delta}}
\umap{α}{03B1}{\ensuremath{\alpha}}
\umap{β}{03B2}{\ensuremath{\beta}}
\umap{²}{00B2}{\ensuremath{^{2}}}
\umap{³}{00B3}{\ensuremath{^{3}}}
\umap{₁}{2081}{\ensuremath{_{1}}}
\umap{₂}{2082}{\ensuremath{_{2}}}
\umap{·}{00B7}{\,\textperiodcentered\,}
\umap{✓}{2713}{\ding{51}}
\umap{✗}{2717}{\ding{55}}
\umap{ }{2009}{\,}   % U+2009 thin space
\umap{ }{00A0}{~}   % U+00A0 no-break space
\umap{‑}{2011}{-}   % U+2011 non-breaking hyphen

\newcommand{\secsquare}{%
  \textcolor{accent}{\rule[0.14em]{0.52em}{0.52em}}\hspace{0.42em}}
\titleformat{\section}
  {\normalfont\large\bfseries\color{headtint}}{\secsquare\thesection}{0.45em}{}
\titlespacing*{\section}{0pt}{16pt plus 4pt minus 2pt}{5pt plus 1pt}
\titleformat{\subsection}
  {\normalfont\normalsize\bfseries\color{headtint}}{\thesubsection}{0.45em}{}
\titlespacing*{\subsection}{0pt}{11pt plus 3pt minus 2pt}{3pt}
\titleformat{\paragraph}[runin]
  {\normalfont\bfseries\color{headtint}}{}{0pt}{}[.]
\titlespacing*{\paragraph}{0pt}{6pt plus 2pt}{0.6em}

\newcommand{\shorttitle}{Testing world models on actions they have never seen}
\renewcommand{\headrulewidth}{0.4pt}
\renewcommand{\headrule}{\hbox to\headwidth{\color{accent!45}\leaders\hrule height \headrulewidth\hfill}}
\input{generated/p3_macros}
\newif\ifjournal\journalfalse
\newif\iftmlr\tmlrfalse

\begin{document}

% ------------------------------------------------------------ title block -
\twocolumn[
  \vspace*{6pt}
  \begin{center}
    {\LARGE\bfseries\color{headtint} Testing World Models on Actions They Have Never Seen:\\[2pt]
     Held-Out Plans, Forecast Accuracy and Price Rules in Retail\par}
    \vspace{10pt}
    {\large Karthik Iyer\par}
    \vspace{3pt}
    {\footnotesize Correspondence:
     \href{mailto:karthik.iyer.au@gmail.com}{karthik.iyer.au@gmail.com}
     \quad ORCID:
     \href{https://orcid.org/0009-0004-0593-1602}{0009-0004-0593-1602}\par}
    \vspace{18pt}
    \begin{minipage}{0.86\textwidth}
      \input{abstract}
    \end{minipage}
    \vspace{16pt}
  \end{center}
]
\thispagestyle{plain}

\input{body}

\end{document}

%% file: generated/p3_macros.tex
\newcommand{\errLLlShape}{0.13}

\newcommand{\errLWm}{1.06}

\newcommand{\regdownLWm}{6.4}

\newcommand{\priceLWm}{9.5}
\newcommand{\errLLgb}{0.93}

\newcommand{\cutsixLLgb}{0.1}
\newcommand{\regdownLLgb}{3.9}

\newcommand{\priceLLgb}{4.2}
\newcommand{\errLLl}{0.82}

\newcommand{\regdownLLl}{6.1}

\newcommand{\priceLLl}{9.5}
\newcommand{\errLLlRr}{0.17}

\newcommand{\errLWmRr}{1.65}

\newcommand{\errLLgbRr}{0.89}

\newcommand{\priceLLgbRr}{3.9}
\newcommand{\errLSplit}{0.17}

\newcommand{\errLShape}{0.39}

\newcommand{\cutsixLShape}{0.5}

\newcommand{\shapeBeatsLlL}{none}

\newcommand{\linkLAllLo}{0.33}
\newcommand{\linkLAllHi}{0.51}

\newcommand{\linkLWmZero}{0.64}

\newcommand{\linkLWmOneHalf}{−0.35}
\newcommand{\linkLWmOneHalfCi}{−0.63 to −0.01}

\newcommand{\errBLlShape}{0.17}

\newcommand{\errBLgbShape}{0.59}

\newcommand{\cutsixBLgbShape}{1.0}
\newcommand{\regdownBLgbShape}{3.5}

\newcommand{\errBWm}{1.39}
\newcommand{\fcBWm}{0.475}

\newcommand{\priceBWm}{9.6}
\newcommand{\errBLgb}{0.89}

\newcommand{\errBLl}{1.22}

\newcommand{\cutsixBLl}{16.8}

\newcommand{\errBLlRr}{0.73}

\newcommand{\errBSplit}{0.71}

\newcommand{\cutsixBSplit}{16.6}

\newcommand{\errBShape}{0.27}
\newcommand{\fcBShape}{0.444}
\newcommand{\cutsixBShape}{1.5}
\newcommand{\regdownBShape}{1.4}

\newcommand{\priceBShape}{0.4}

\newcommand{\shapeBeatsLlB}{one}

\newcommand{\tlLlShapeB}{0.14}
\newcommand{\tlShapeB}{0.23}
\newcommand{\tlLgbB}{0.82}
\newcommand{\tlWmB}{1.37}
\newcommand{\linkBAllLo}{0.32}
\newcommand{\linkBAllHi}{0.66}

\newcommand{\linkBWmZero}{0.43}

\newcommand{\linkBWmOne}{−0.69}
\newcommand{\linkBWmOneCi}{−0.79 to −0.37}

\newcommand{\cutsixKLlShape}{3.4}

\newcommand{\errKWm}{0.92}

\newcommand{\errKShape}{0.30}

\newcommand{\cutsixKShape}{1.6}

\newcommand{\linkAllMin}{0.32}
\newcommand{\linkAllMax}{0.66}
\newcommand{\nFits}{2,853}
\newcommand{\errLActLo}{1.05}
\newcommand{\errLActHi}{1.06}
\newcommand{\errBActLo}{1.34}
\newcommand{\errBActHi}{1.39}

\newcommand{\hhBaseKWmLlWon}{36}
\newcommand{\hhBaseKWmLlN}{40}
\newcommand{\hhBaseKWmLlP}{below 0.001}
\newcommand{\hhBaseKWmLlDiff}{−0.12}
\newcommand{\hhBaseKWmLlCi}{−0.18 to −0.09}

\newcommand{\hhBaseKWmLgbWon}{38}
\newcommand{\hhBaseKWmLgbN}{40}

\newcommand{\hhMedKWmLlWon}{12}
\newcommand{\hhMedKWmLlN}{12}

\newcommand{\hhBestBWmLlWon}{1}
\newcommand{\hhBestBWmLlN}{12}

\newcommand{\hhBestBWmLlA}{0.20}
\newcommand{\hhBestBWmLlB}{0.12}

\newcommand{\hhBestKWmLlWon}{12}
\newcommand{\hhBestKWmLlN}{12}

\newcommand{\reCPureLgb}{0.50}

\newcommand{\reCPureWm}{0.45}

\newcommand{\reCPureWmShape}{0.44}

\newcommand{\reFDeepLgb}{0.41}

\newcommand{\reFDeepWm}{0.39}

\newcommand{\reFDeepWmShape}{0.43}
\newcommand{\reFPureLgb}{0.53}

\newcommand{\reFPureLlShape}{1.26}
\newcommand{\reFPureWm}{0.63}

\newcommand{\reFPureWmShape}{0.56}

\newcommand{\reTPureLgb}{0.43}

\newcommand{\reTPureWm}{0.49}

\newcommand{\reTPureWmShape}{0.47}

\newcommand{\reMMoveLl}{0.52}

\newcommand{\reMMoveWm}{0.55}

\newcommand{\rnFPure}{6,288}
\newcommand{\roFPure}{eight}

\newcommand{\rnCPure}{324}

\newcommand{\rnFMove}{1,115}

\newcommand{\rpCPureWmShapeVWm}{−0.021}
\newcommand{\rpCPureWmShapeVWmCi}{−0.050 to 0.012}

\newcommand{\rpCPureWmShapeVLgb}{−0.065}
\newcommand{\rpCPureWmShapeVLgbCi}{−0.127 to 0.000}
\newcommand{\rpCPureWmShapeVLgbWon}{four}
\newcommand{\rpCPureWmShapeVLgbOf}{five}

\newcommand{\rpFMoveWmShapeVWm}{0.078}
\newcommand{\rpFMoveWmShapeVWmCi}{0.048 to 0.108}

\newcommand{\rpFPureWmShapeVWm}{−0.113}
\newcommand{\rpFPureWmShapeVWmCi}{−0.137 to −0.088}
\newcommand{\rpFPureWmShapeVWmWon}{eight}
\newcommand{\rpFPureWmShapeVWmOf}{eight}

\newcommand{\rpFPureWmShapeVLlShape}{−0.681}
\newcommand{\rpFPureWmShapeVLlShapeCi}{−0.746 to −0.613}

\newcommand{\rpTPureWmShapeVWm}{−0.048}
\newcommand{\rpTPureWmShapeVWmCi}{−0.056 to −0.041}
\newcommand{\rpTPureWmShapeVWmWon}{seven}
\newcommand{\rpTPureWmShapeVWmOf}{eight}

\newcommand{\dsLinkLo}{0.42}
\newcommand{\dsLinkHi}{0.71}
\newcommand{\dsWmLinkHigh}{−0.40}
\newcommand{\dsWmLinkHighCi}{−0.66 to −0.06}
\newcommand{\dsWmZero}{0.75}

\newcommand{\dsWmRatio}{−2.19}
\newcommand{\dsLgbRatio}{−2.96}
\newcommand{\dsShapeRatio}{0.06}
\newcommand{\dsLlShapeRatio}{0.00}
\newcommand{\dsLgbShapeRatio}{−0.64}
\newcommand{\dsFits}{408}
\newcommand{\dsExpShapeErr}{1.91}
\newcommand{\dsShapeErr}{2.86}
\newcommand{\dsExpRatioLo}{0.12}
\newcommand{\dsExpRatioHi}{0.32}

\newcommand{\dsExpVsLlWon}{ten}
\newcommand{\dsExpVsLlP}{0.039}
\newcommand{\dsExpVsLgbWon}{ten}

\newcommand{\nsKPlainA}{0.91}
\newcommand{\nsKPlainB}{0.93}

\newcommand{\nsBShapeA}{0.27}
\newcommand{\nsBShapeB}{0.27}
\newcommand{\nsKShapeWon}{14}
\newcommand{\nsKShapeN}{40}
\newcommand{\nsKShapeP}{0.08}

\newcommand{\nsKShapeA}{0.29}
\newcommand{\nsKShapeB}{0.28}
\newcommand{\nsKLlWon}{38}
\newcommand{\nsKLlN}{40}

%% file: abstract.tex
{\small
\textbf{Abstract.} A world model is trusted to answer ``what if'': how its world responds to an action it has never
seen. We give a test of that trust for any action-conditioned world model, in three parts: actions held out of
training by design and scored against a known answer; whether forecast accuracy predicts accuracy on those actions;
and the same structure given to simpler models, to separate what the structure does from what the world model
learns. We run it in retail, where the action is a price and promotion plan: models are trained on the plans
merchants make and asked to forecast plans held out by design: deeper cuts than any in training, lasting regular-price moves, promotions moved to quiet weeks and no
plan at all. Across \nFits{} fitted models we find four things. First, forecast accuracy is a fair guide between
kinds of model (rank correlation with error on new plans \linkAllMin{} to \linkAllMax{}) and a poor guide between
versions of one world model: where merchants plan into high-demand weeks the link reverses, down to \linkBWmOne{}
(interval \linkBWmOneCi{}). Second, training the world model to learn the effect of actions (JEPA pretraining, a
latent consistency check, adaptation at the plan date) leaves its error on new plans unchanged (\errLActLo{} to
\errLActHi{} against \errLWm{}). Third, in the generated worlds what improves it is the recorded regular price and three price rules: sales may
not fall as a cut deepens, the gain from deeper cuts may flatten, and a lasting regular price acts through its own
elasticity. Given the same rules, a log-linear model is best when its curve can match the truth (error \errBLlShape{}
against \errBShape{} for the world model with the rules and \errBLgbShape{} for LightGBM with the rules); in a world
whose curve lies outside the rules, the world model with the rules is better than the simpler models we tested
(ahead of the log-linear model with the rules in \hhBaseKWmLlWon{} of \hhBaseKWmLlN{} worlds, sign test $p$
\hhBaseKWmLlP{}, and in \hhMedKWmLlWon{} of \hhMedKWmLlN{} when every model is tuned over its versions), and does as well without its recurrent cell. On real data
(Breakfast at the Frat, The Complete Journey and M5), where only the rarest plans that were run can be scored, the
rules lower the world model's error on pure price cuts deeper than any it saw (\reFPureWm{} to \reFPureWmShape{} on
the Frat; \reCPureWm{} to \reCPureWmShape{}, within chance, on The Complete Journey), making it the most accurate model on one data set and close to LightGBM on the
other, while neither the rules nor the recorded regular price help with lasting price moves. Fourth, a small share of random test plans helps every model a little. In a generated dosing world, a stress test in a second domain where doctors
dose sicker patients more, the first finding repeats; the dose rules fix the sign of a higher dose's effect but not its
size, and adding the drug level built up from recent doses lowers the world model's error (\dsShapeErr{} to
\dsExpShapeErr{}) though it still predicts only \dsExpRatioLo{} to \dsExpRatioHi{} of the true gain where doses
follow severity. The test carries to any domain where a world model is asked about actions outside its history.
\par}

%% file: body.tex
% Paper 3 body. Every number is a macro from generated/p3_macros.tex or a table from generated/, written by
% figs/make_figures.py from results/paper3.

\section{Introduction}\label{sec:intro}

A world model earns its keep on actions not yet taken: a robot's next grasp, a route not yet driven, a price not yet
set. Its accuracy on the data it was fitted to says little about this, because logged actions follow a policy, and
the actions that matter for planning are the ones the policy rarely took~\cite{levine2020}. We give a test of a world
model on actions it has never seen, in three parts that apply to any action-conditioned world model: (1) actions held
out of training by design and scored against a known answer; (2) whether forecast accuracy, the usual way to choose
a model, predicts accuracy on those actions; (3) the same structure given to simpler models, so that what the
structure does is separated from what the world model learns. Each part is an established idea (held-out actions, evaluation off the
logging policy, structured baselines); what is new is their combination into one test and what it finds.

We run the test in retail, where the action is a price and promotion plan and the stakes are concrete: a merchant
asks what a deeper cut, a new regular price or a promotion in a quiet week would bring. Our forecasting
study~\cite{rwm} compared world models with trees and foundation models on the plans merchants made; it cannot say
whether a model is right about plans nobody made. In generated stores where the expected sales of any plan are
known~\cite{paper2}, we train models on the plans merchants make and ask them to forecast plans held out by design,
then check the result on the rarest plans in three real data sets. We find:

\begin{itemize}\setlength{\itemsep}{1pt}
\item \textbf{Forecast accuracy is a fair guide between kinds of model, and a poor one between versions of a world
model.} Across all 25 versions in each world, better forecasters simulate new plans better (\linkAllMin{} to
\linkAllMax{}, every interval above zero). Within the world model's versions, where plans follow demand, the link
reverses (\linkBWmOne{}, \linkBWmOneCi{}).
\item \textbf{Training the world model to learn actions does not fix it.} JEPA pretraining, a latent consistency
check, a check seven weeks ahead and adaptation at the plan date move its error on new plans by at most 0.05.
\item \textbf{The recorded regular price and three price rules improve it.} Without the regular price every model reads
a lasting price move as a temporary cut. With it, and with rules that keep sales from falling as a cut deepens, let
the gain from deeper cuts flatten and give a lasting regular price its own elasticity, the world model's error falls
from \errBWm{} to \errBShape{} in a world whose response bends.
\item \textbf{The rules do most of the work; against the simpler models we tested, the world model adds when the
truth lies outside them.} Given the same
rules, a log-linear model does better still where its curve can match the truth (\errBLlShape{} in the bent world),
and LightGBM worse (\errBLgbShape{}). In a world whose curve neither rule set can draw, the world model with the rules
is ahead of the log-linear model with the rules in \hhBaseKWmLlWon{} of \hhBaseKWmLlN{} worlds and of LightGBM with
the rules in \hhBaseKWmLgbWon{}, and ahead of the log-linear model with the rules in every one of twelve worlds when
each model is tuned over its versions.
\item \textbf{On real data the rules help the world model with deep cuts it has not seen, not with lasting price
moves.} On pure price cuts (no circular, no display) deeper than any in training, the rules lower the world model's
error on Breakfast at the Frat (\reFPureWm{} to \reFPureWmShape{}, lower at \rpFPureWmShapeVWmWon{} of
\rpFPureWmShapeVWmOf{} origins) and on The Complete Journey, where it is the most accurate model (\reCPureWmShape{}
against \reCPureLgb{} for LightGBM); on the Frat LightGBM stays slightly ahead (\reFPureLgb{}). On lasting
regular-price moves neither the rules nor the recorded regular price help.
\end{itemize}

The rest of the paper sets out the stores (\S\ref{sec:data}), the models and remedies (\S\ref{sec:method}), the
setup (\S\ref{sec:setup}) and the results (\S\ref{sec:results}). The plan for this study was fixed before any run,
and each of the nine amendments was written before the runs it added (Appendix~\ref{app:plan}).

\section{Related work}\label{sec:related}

\paragraph{World models} Learned models of how an environment responds to actions are used to plan in games and
control~\cite{worldmodels,planet,dreamerv3,tdmpc2}. Joint-embedding predictive architectures learn such models by
predicting in a latent space rather than in the raw signal~\cite{lecun2022,ijepa,vjepa2,leworldmodel}; recent work
separates the effect of the action from the drift of the world~\cite{dwm} and trains the model to tell actions
apart~\cite{adwm}. Our molecular work found that predictive pretraining transfers to scaffolds not seen in
training~\cite{mjepa}. We test whether these training choices make a retail world model right about plans outside
its history. They do not; structure in the price response does, and then mostly through the structure itself.

\paragraph{Retail promotion and price models} Log-linear models of sales on price, promotion and display have long
been the retailer's tool for promotion planning~\cite{promocast,vanheerde2000,vanheerde2003}, with well-studied
elasticities~\cite{hoch1995,bijmolt2005}. Machine-learning forecasters now lead on accuracy~\cite{m5accuracy,
fildes2022,ma2016}. We use the log-linear model as the reference a world model has to beat on plans, and show where
its fixed form fails.

\paragraph{Extrapolation and structure} Models fitted to logged decisions are unreliable for actions the logging
policy rarely took~\cite{levine2020,pearl2009}, and models with equal held-out accuracy can behave very differently
off the data~\cite{damour2022}. Monotone networks~\cite{sill1998,you2017} and physics-informed models~\cite{
karniadakis2021} build known structure into a learned model so it extrapolates sensibly. We bring that idea to the
price response of a retail world model, and give the same structure to the simpler models so its value can be
separated from the world model's.

\paragraph{Our own studies} The forecasting study~\cite{rwm} measures accuracy on observed plans on four public data
sets; the explanation audit~\cite{paper2} tests whether a model's stated lift is right. This paper reuses no result
of either: its worlds use new seeds, three new world shapes and new held-out plans, and its measures are new.

\section{Generated stores with known answers}\label{sec:data}

Each world has 18 stores in three price zones and 20 items over 260 weeks. Weekly sales are Poisson around an
expected level set by the item's size, the season, a persistent demand shock and the plan: the price (a temporary
cut below the regular price), a promotion, a display, the week after a planned week, and the regular price itself.
The truth for any plan is the expected level the generator gives it, so the effect of a plan never run is known
exactly.

\paragraph{Planning into demand} Merchants plan cuts and promotions into weeks they expect to be strong. A
planning strength of 0 places plans at random; strengths 0.5, 1 and 1.5 tie them more closely to the demand shock,
so a model that reads the plan as the cause of high sales overstates its effect.

\paragraph{Three shapes of world} In the \emph{log-linear world} a cut of depth $c$ multiplies sales by
$(1-c)^{-2.5}$ and the regular price acts with elasticity $-0.5$; this is the form of the log-linear model, which
therefore matches the truth by construction. In the \emph{bent world} the gain from a cut flattens,
$1 + 1.015\,(1 - e^{-c/0.15})$ (equal to the log-linear world at a 20\% cut), the regular-price response is linear
in the price ratio, and a promotion with a display adds less than both alone. In the \emph{kinked world} a cut adds
nothing up to 10\%, then sales rise in a straight line to 1.6 times at 40\% and stay there; the rest is as in the bent
world. Neither the log-linear form nor the price rules below can draw this curve, so no model has its shape built in.
Each shape is generated at four planning strengths and three seeds: 36 worlds.

\paragraph{Test and learn} A further 24 worlds (log-linear and bent) add random test plans: 5\% of training item-weeks get a cut of 10\%
to 40\% (the merchants' own range), a promotion and a display, each with probability one half, placed apart from
demand. The tests come from their own random stream, so everything else in the world is unchanged. The cuts stay
inside the range merchants use, so the deeper cuts held out below remain unseen.

\paragraph{Real data (added)} Breakfast at the Frat~\cite{frat}: 77 stores, 55 items, 156 weeks, with each week's
regular price recorded; and M5~\cite{m5accuracy}, the FOODS\_3 department in three stores, summed to weeks, with a
regular price constructed from the prices (a price held for four weeks or more becomes the regular price); and The
Complete Journey~\cite{completejourney}: purchases by 2,500 households summed to product by week over all stores
(102 weeks, the 756 food and household products bought in at least 95 weeks), with the shelf price before the
loyalty discount as the recorded regular price and the weeks each product was in the mailer or on display. All are
scored on the rarest plans that were run (\S\ref{sec:setup}).

\section{Models and remedies}\label{sec:method}

\paragraph{World model} The state model of our forecasting study~\cite{rwm}: an encoder reads each item's last 13
or 26 weeks of sales and plan into a state, a recurrent cell rolls the state forward under the plan, and a readout
gives expected sales as a baseline times a lift. Twelve versions per world vary the width (16, 32, 64), training
steps (500, 2{,}000) and history (13, 26 weeks).

\paragraph{LightGBM} Gradient-boosted trees~\cite{lightgbm} with the plan as features, one model per horizon week.
Twelve versions vary the leaves (7, 31, 127), rounds (100, 500) and minimum rows per leaf (20, 100).

\paragraph{Log-linear model} Poisson regression of sales on the series, season, log price over the regular price,
log regular price over the series' mean, promotion, display, both together and the week after a planned week. The
regular price is the highest price of the last 12 weeks, as for the other models.

\paragraph{Remedies added after the first results} Each changes one thing and is given to every model it applies
to on the same data.
\begin{enumerate}\setlength{\itemsep}{1pt}
\item \emph{Learning actions}: JEPA pretraining (500 steps on the latent loss first), the latent consistency check,
the check seven weeks ahead, and adaptation at the plan date.
\item \emph{The recorded regular price}: retailers know whether a price is a cut or a new regular price. Every
model is given it (versions marked ``+ recorded regular price'').
\item \emph{Split the job}: the forecaster gives the week at the current regular price with nothing planned; the
log-linear model with the recorded regular price gives the change the plan makes.
\item \emph{Price rules in the world model}: the plan's price reaches the world model only through
\begin{equation}\label{eq:shape}
\log m = a_1\bigl(1 - e^{-c/b}\bigr) - a_2 \log(1-c) - \gamma \log\frac{r}{\bar p},
\end{equation}
where $c$ is the cut (capped at 95\%), $r$ the recorded regular price, $\bar p$ the item's usual price, and
$a_1, a_2, b, \gamma > 0$ are learned per item and store from their labels. Sales cannot fall as a cut deepens. The
first term flattens at $a_1$; the second grows without limit as the cut nears 100\% and is held finite only by the
cap on the cut, so the model learns from the data how far the gain keeps rising. A lasting regular price acts
through its own elasticity. The rest of the network sees
neither price, so it cannot learn a price effect that breaks these rules.
\item \emph{The same rules in the simpler models}, so their value can be told apart from the world model's: the
log-linear model with the recorded regular price and Equation~\eqref{eq:shape} in place of log price ($b$ on a grid of
0.05, 0.1, 0.2 and 0.4; coefficients of the cut at or above zero, of the regular price at or below zero; Poisson
maximum likelihood with those bounds); and LightGBM with the recorded regular price and monotone constraints, so
predicted sales cannot rise with any price feature or with the regular price. Trees stay flat beyond the deepest cut
seen, so LightGBM's gain flattens by itself.
\item \emph{Test and learn}: every model refitted on the worlds with random test plans.
\end{enumerate}

\section{Experimental setup}\label{sec:setup}

Models are fitted on weeks 1 to 252 and forecast weeks 253 to 260 from the same origin, under the merchants' own
plan and under eight held-out plans: every planned cut deepened to 50\% and to 60\% (training cuts run from 10\% to
40\%); the regular price moved 10\% and 5\% down and up for the whole horizon; promotions and displays moved to each
series' lowest-demand weeks; and no plan. On the real data, models are fitted before each of several origins (eight on the Frat, five on The
Complete Journey, four on M5) and forecast the 8 weeks after each. On the Frat and The Complete Journey every
item-week with a cut deeper than 30\% off the regular price (about the 95th and 90th percentiles) is removed from
training; on the Frat we repeat this at 20\%. The held-out weeks are scored on cuts deeper than the threshold, on
the pure price cuts among them (no circular and no display, so price is the only lever that changed), on lasting
regular-price moves (on the Frat, 10\% or more, steady for four weeks before the origin and through every held-out
week; on M5, 5\% or more), and on the rest.

\paragraph{Measures} \emph{Forecast accuracy}: item-by-store RMSSE on the merchants' own plan. \emph{Error on new
plans}: for each held-out plan, the predicted change in sales (held-out plan minus own plan, summed) over the true
change; the error is $|\log|$ of that ratio, 0 when the change is right, capped at a twentyfold miss, which a wrong
sign also counts as; we report the mean over the eight plans. \emph{Decisions}: the revenue given up by the
regular-price move (of $-10\%, -5\%, 0, +5\%, +10\%$) the model picks; the true lift missed by switching on the
promotions with the largest predicted lift in half the item-weeks; and the error of ordering stock to the forecast.

\paragraph{The test} Within each world, the rank correlation over the 25 versions between forecast error and each
measure, averaged over three seeds, with a 95\% interval from 2{,}000 bootstrap draws of versions. The versions share
each world's data, so this interval shows how the correlation varies over versions on fixed data, not over
independent replications; the three seeds at each planning strength are the replications. The same correlation
within each model class was added after the first results. Comparisons between models (added) take the world as the
unit: the number of worlds in which one model has the lower error, with a two-sided sign test, and the median paired
difference with a 95\% interval from a bootstrap over worlds.

\section{Results}\label{sec:results}

\subsection{Better forecasting is a fair guide between kinds of model, not between versions}

\begin{figure*}[t]
\centering
\includegraphics[width=\textwidth]{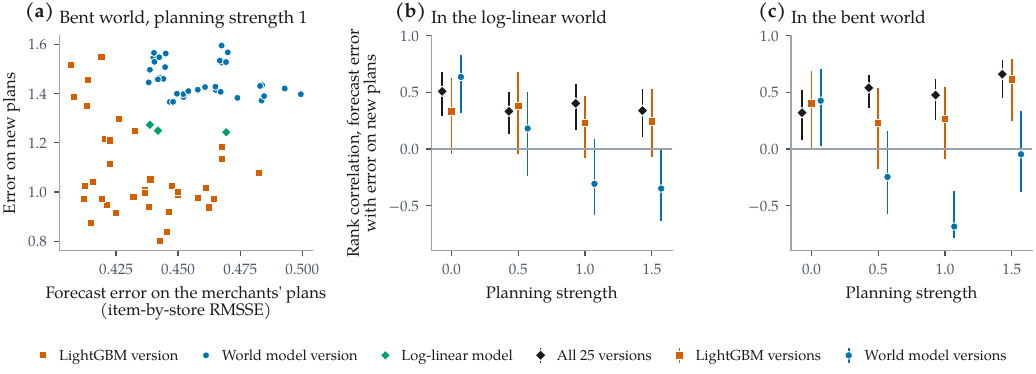}
\caption{\textbf{Forecast accuracy and error on new plans.} (a) Every version in the three bent worlds at planning
strength 1: forecast error on the merchants' plans against error on the held-out plans. (b, c) Rank correlation of
the two within each world, averaged over three seeds, with 95\% bootstrap intervals; all versions (the planned test)
and within each model class (added).}
\label{fig:link}
\end{figure*}

Across all 25 versions, the models that forecast the merchants' plans better also simulate new plans better, in
every world (Table~\ref{tab:link}; Figure~\ref{fig:link}b,c): \linkLAllLo{} to \linkLAllHi{} in the log-linear
world and \linkBAllLo{} to \linkBAllHi{} in the bent world, every interval above zero. The link comes mostly from
the gap between kinds of model: LightGBM forecasts and simulates better than the plain world model
(Figure~\ref{fig:link}a).

Within the world model's own versions the link depends on how merchants plan. With plans placed at random it holds
(\linkLWmZero{} and \linkBWmZero{}). Where plans follow demand it reverses: \linkBWmOne{} (\linkBWmOneCi{}) in the
bent world and \linkLWmOneHalf{} (\linkLWmOneHalfCi{}) in the log-linear world at the strongest planning. The
versions that forecast best are those that learn most from the merchants' plans, and where plans follow demand
those plans teach the wrong effect. Choosing a world model version on forecast accuracy can choose the worst
simulator. Stock error tracks forecast accuracy closely in every class (Table~\ref{tab:link}), as ordering to a
forecast should.

\input{generated/tab_link}

\subsection{Where the models go wrong}

\begin{figure*}[t]
\centering
\includegraphics[width=\textwidth]{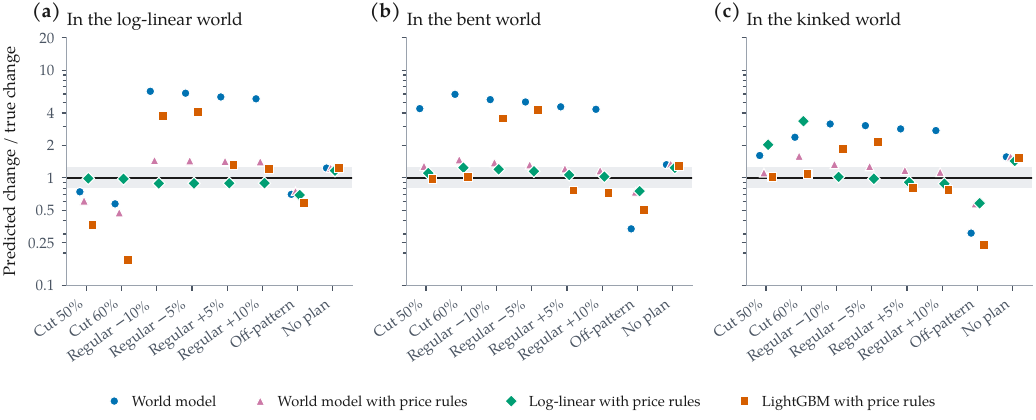}
\caption{\textbf{Predicted change in sales over the true change, for each held-out plan.} Medians over the 12 worlds
of each shape (planning strengths and seeds). 1 is right; the band marks 0.8 to 1.25. Log scale.}
\label{fig:ratio}
\end{figure*}

Figure~\ref{fig:ratio} shows the predicted change over the true change for each held-out plan. Three failures stand
out. Every model without the recorded regular price reads a lasting regular-price move as a temporary cut: the
plain world model predicts \regdownLWm{} times the true gain from a 10\% lower regular price, LightGBM \regdownLLgb{}
times and the log-linear model \regdownLLl{} times, and each loses \priceLWm\%, \priceLLgb\% and \priceLLl\% of
revenue on the regular-price choice in the log-linear world. LightGBM cannot extend a response past the prices it
saw: it predicts \cutsixLLgb{} of the true gain from a 60\% cut. And the log-linear model extends its curve without
limit: in the bent world it predicts \cutsixBLl{} times the true gain from a 60\% cut.

\subsection{Training the world model to learn actions does not help}

\begin{figure*}[t]
\centering
\includegraphics[width=\textwidth]{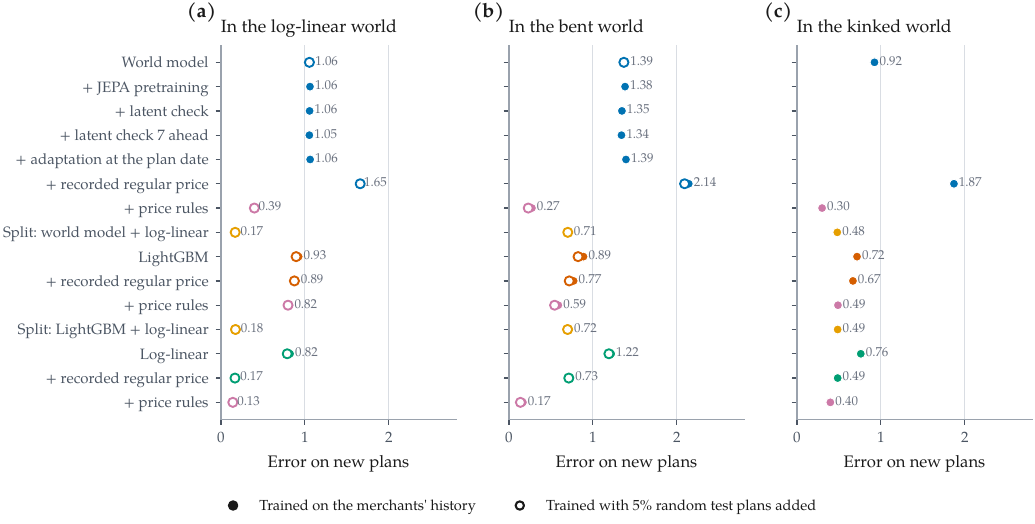}
\caption{\textbf{Error on new plans for every model and remedy} (0 means the predicted change in sales is right).
Medians over the 12 worlds of each shape. Filled: trained on the merchants' history; open: with 5\% random test
plans. Rows after the first in each block, and the split rows, were added after the first results.}
\label{fig:remedies}
\end{figure*}

The four changes meant to make the world model learn the effect of actions leave its error on new plans where it
was: \errLActLo{} to \errLActHi{} against \errLWm{} in the log-linear world and \errBActLo{} to \errBActHi{}
against \errBWm{} in the bent world (Table~\ref{tab:remedies}, Figure~\ref{fig:remedies}). They change how the
state is learned; they do not change what the history contains, and the history holds no lasting price move
labelled as one and no cut deeper than 40\%.

\subsection{What works}

\paragraph{The recorded regular price} Given the recorded regular price, the log-linear model's error falls from
\errLLl{} to \errLLlRr{} in the log-linear world and from \errBLl{} to \errBLlRr{} in the bent world, and its
regular-price choice loses nothing. LightGBM gains little (\errLLgb{} to \errLLgbRr{}): trees cannot extend a
response beyond the regular prices seen. The plain world model given the regular price picks the right price move
but predicts almost no change in sales from it, so its error rises (\errLWmRr{}); a free network has no reason to
give the new input a sensible effect.

\paragraph{Split the job} The world model for the week with nothing planned and the log-linear model for the change
lowers the error to \errLSplit{} in the log-linear world, as good as the log-linear model alone, and to \errBSplit{}
in the bent world. It inherits the log-linear model's form, so in the bent world it overstates a 60\% cut
\cutsixBSplit{} times.

\paragraph{Price rules in the world model} Building the rules of Equation~\eqref{eq:shape} into the world model
lowers its error on new plans from \errBWm{} to \errBShape{} in the bent world and from \errKWm{} to \errKShape{} in
the kinked world. It predicts \cutsixBShape{} times the true gain from a 60\% cut in the bent world and
\regdownBShape{} times that of a 10\% lower regular price, loses \priceBShape\% on the price choice, and forecasts the
merchants' plans as well as before (\fcBShape{} against \fcBWm{}). In the log-linear world it learns too small a
logarithmic term and understates the steep gain of a 60\% cut (\cutsixLShape{} of the truth; error \errLShape{}).

\paragraph{Test and learn} Random test plans inside the merchants' range help every model a little: the world model
with the rules from \errBShape{} to \tlShapeB{}, the log-linear model with the rules from \errBLlShape{} to
\tlLlShapeB{} and LightGBM from \errBLgb{} to \tlLgbB{} in the bent world; the plain world model
barely moves (\errBWm{} to \tlWmB{}). Tests break the link between plan and demand but cannot show a cut deeper
than the deepest tested; a model still needs a sensible shape to go beyond it.

\input{generated/tab_remedies}

\subsection{The rules or the world model?}

The world model with the rules might win only because its rules match the worlds. Giving the same rules to the
simpler models separates the two (Table~\ref{tab:remedies}, Figure~\ref{fig:remedies}). Where the rules' curve can
match the truth, the log-linear model with the rules is best: \errLLlShape{} in the log-linear world and
\errBLlShape{} in the bent world, against \errLShape{} and \errBShape{} for the world model with the rules, which is ahead in
only \shapeBeatsLlB{} of the twelve bent worlds and \shapeBeatsLlL{} of the log-linear ones. LightGBM with the rules gets deep cuts right in the bent world (\cutsixBLgbShape{} times the truth
at 60\%) but still cannot carry a lasting regular-price move beyond the prices it saw (\regdownBLgbShape{} times the
truth at $-10\%$), and its error is \errBLgbShape{}.

In the kinked world, whose curve neither rule set can draw, the order turns. The world model with the rules is
ahead of the log-linear model with the rules in \hhBaseKWmLlWon{} of \hhBaseKWmLlN{} worlds (sign test $p$
\hhBaseKWmLlP{}; median difference in error \hhBaseKWmLlDiff{}, interval \hhBaseKWmLlCi{}) and of LightGBM with the
rules in \hhBaseKWmLgbWon{} of \hhBaseKWmLgbN{} (Table~\ref{tab:h2h}). The log-linear model has to bend its fixed
curve to the kink and overstates a 60\% cut \cutsixKLlShape{} times; the world model, whose rules set only the
direction and whose network learns the rest, overstates it \cutsixKShape{} times.

\paragraph{Equal tuning} Tuned over its own versions (12 for the world model and LightGBM, six for the log-linear
model), each model keeps its place: in the kinked world the world model with the rules is ahead of the log-linear
model with the rules in \hhMedKWmLlWon{} of \hhMedKWmLlN{} worlds at the median version and \hhBestKWmLlWon{} of
\hhBestKWmLlN{} at the best; in the bent world the world model with the rules is ahead of the log-linear model in only
\hhBestBWmLlWon{} of \hhBestBWmLlN{} worlds at the best version (errors \hhBestBWmLlA{} against \hhBestBWmLlB{}).

\paragraph{Without the recurrent cell (added)} With the cell that rolls the state forward under the plan removed, the
world model with the rules scores \nsKShapeA{} against \nsKShapeB{} in the kinked world (ahead in \nsKShapeWon{} of
\nsKShapeN{} worlds, $p$ \nsKShapeP{}) and \nsBShapeA{} against \nsBShapeB{} in the bent world, and is still ahead of
the log-linear model with the rules in \nsKLlWon{} of \nsKLlN{} kinked worlds; without the rules, \nsKPlainA{} against
\nsKPlainB{}. Removing the cell did not materially change performance here, so the advantage cannot be attributed to the recurrent state.

\input{generated/tab_h2h}

\subsection{Real data}

The generated worlds score plans never run; real data cannot. The nearest real test is the rarest plans that were
run (Table~\ref{tab:real}), and the cleanest of these are pure price cuts deeper than any the models saw, where
price is the only lever that changed.

\paragraph{Pure price cuts} The rules lower the world model's error in all three tests of pure price cuts, clearly on
the Frat and by a margin within chance on The Complete Journey. On the Frat with cuts over
30\% held out (\rnFPure{} item-weeks over \roFPure{} origins) it falls from \reFPureWm{} to \reFPureWmShape{}
(difference pooled over origins \rpFPureWmShapeVWm{}, interval \rpFPureWmShapeVWmCi{} from a bootstrap over series; lower at
\rpFPureWmShapeVWmWon{} of \rpFPureWmShapeVWmOf{} origins); with cuts over 20\% held out, from \reTPureWm{} to
\reTPureWmShape{} (\rpTPureWmShapeVWm{}, \rpTPureWmShapeVWmCi{}; \rpTPureWmShapeVWmWon{} of
\rpTPureWmShapeVWmOf{}); on The Complete Journey (\rnCPure{} item-weeks), from \reCPureWm{} to \reCPureWmShape{}
(\rpCPureWmShapeVWm{}, \rpCPureWmShapeVWmCi{}), where the world model with the rules is the most accurate of the
nine models, ahead of LightGBM (difference \rpCPureWmShapeVLgb{}, \rpCPureWmShapeVLgbCi{}; \rpCPureWmShapeVLgbWon{} of
\rpCPureWmShapeVLgbOf{} origins). On the Frat, plain LightGBM stays slightly ahead (\reFPureLgb{} and
\reTPureLgb{}). The log-linear model with the rules is the least reliable on the Frat's deepest pure cuts
(\reFPureLlShape{}): its curve, fitted to cuts up to 30\%, runs on past the data, as in the bent and kinked worlds,
and the world model with the rules is far ahead of it (difference \rpFPureWmShapeVLlShape{},
\rpFPureWmShapeVLlShapeCi{}).

\paragraph{Deep cuts with a circular or display} When every deep cut is counted, most come with a circular or a
display, which every model has seen in training, and the rules do not help: the world model's error rises (Frat:
\reFDeepWm{} without, \reFDeepWmShape{} with; LightGBM \reFDeepLgb{}).

\paragraph{Lasting regular-price moves} Neither the rules nor the recorded regular price help. On the Frat's clean
moves (\rnFMove{} item-weeks) they raise the world model's error by \rpFMoveWmShapeVWm{}
(\rpFMoveWmShapeVWmCi{}); on M5 every model lies between \reMMoveLl{} and \reMMoveWm{}. These moves are few, and the
regular prices are recorded loosely on the Frat and constructed on M5.

So the real data confirm one half of the generated result: rules about how price can act help a world model on deep
cuts it has not seen. They do not yet confirm the other half, the recorded regular price for lasting moves.

\input{generated/tab_real}

\subsection{A second domain: dosing}

As a stress test of the evaluation in a second domain, not a clinical model, we ran it on a generated dosing world,
reusing the retail models with the dose in the price columns (\dsFits{} fits; Appendix~\ref{app:dosing}): patients whose severity lowers a daily response, a drug that accumulates and
raises it along a saturating curve, and doctors who dose more when a patient is sicker, the clinical form of
planning into demand. Higher doses than any given, a steady daily dose, doses on the mildest days and no dose are
held out. The first finding repeats: across all versions, better forecasters simulate new doses better
(\dsLinkLo{} to \dsLinkHi{}), and within the world model's versions the link reverses where doses follow severity
(\dsWmLinkHigh{}, \dsWmLinkHighCi{}). The second finding does not: where doses follow severity every model without
the rules gets the sign of a higher dose wrong (the world model predicts \dsWmRatio{} times the true change for a
60\% dose, LightGBM \dsLgbRatio{}), and the rules fix the sign but not the size for the world model and the log-linear model (\dsShapeRatio{} and
\dsLlShapeRatio{}), while LightGBM with the rules still gets the sign wrong (\dsLgbShapeRatio{}). The drug acts through the level built up over days, which the rules do not
describe. Given that level in place of the day's dose (added), as the retail models were given the recorded regular
price, the world model with the rules lowers its error from \dsShapeErr{} to \dsExpShapeErr{}, in all twelve worlds,
and is ahead of the log-linear and LightGBM models with the rules in \dsExpVsLlWon{} and \dsExpVsLgbWon{} of twelve
($p$ \dsExpVsLlP{}); where doses follow severity it still predicts only \dsExpRatioLo{} to \dsExpRatioHi{} of the
true gain from a 60\% dose. With doses at random every model is close (\dsWmZero{} for the plain world model).

\subsection{Decisions}

The decision losses follow the errors on new plans (Table~\ref{tab:remedies}). The regular-price choice is where
models differ most: up to \priceBWm\% of revenue given up without the recorded regular price, and under 1\% for the
log-linear, split and rule models that use it; LightGBM still gives up \priceLLgbRr\% with it. The
promotion choice costs every model 2\% or less, because choosing which weeks to promote needs only the ranking of
lifts, which every model gets roughly right. Stock error is lowest for the log-linear model with the rules in the log-linear and bent worlds and for the world
model with the rules in the kinked world.

\section{Discussion}\label{sec:discussion}

The world model fails on new plans for the reason any learner fails off its data: the merchants' history never
shows a deep cut or a lasting price move labelled as one, and where plans follow demand it shows the wrong effect.
No training objective recovers what the data do not contain. Two things help. The first is information the retailer
already has: the recorded regular price. The second is knowledge of how price can act: more discount never loses
sales, the gain from deeper cuts may flatten, a regular price is not a cut.

The controls show where the gain comes from. The rules do most of the work, and a log-linear model given them is
the best simulator when its curve can match the truth. The world model's own learning adds when the true curve is
of a shape nobody wrote down: its rules fix only the direction and the split between cuts and regular prices, and
its network learns the rest. We show this against the simpler models we tested, in one such world; it does not
prove the same for every unknown shape.

In the generated worlds this gives four rules for a retailer; the real data confirm the second for deep price cuts
and do not yet confirm the first. Record the regular price and give it to the model. Give the model price rules
rather than leaving a free model to find them. Use a simple model with the rules when the shape of the price response
is known, and test a flexible model with the rules against it when it is not. Choose between versions on held-out plans, generated or
tested, and not on forecast accuracy, which can pick the worst simulator where plans follow demand.

\section{Limitations and threats to validity}\label{sec:limits}

All three worlds are generated, with price responses of our choosing. That is what lets us score plans never run,
since the answer is known. The real-data check scores sales that happened, not plans never run, and lasting regular-price
moves are few in public data; a real response can
differ from all three shapes, and real data have problems the worlds leave out (stock-outs, competitors, store
changes). The rules in Equation~\eqref{eq:shape} assume
sales never fall as a cut deepens; shoppers who read a very deep cut as a sign of poor quality, or shelves emptied
by a deep cut, would break them. The 40 kinked worlds differ in seed and planning strength but share one design, so
winning in 36 of them shows an advantage within that design, not across kinds of retail environment. The
held-out plans move one lever at a time. Each world has 360 series and eight forecast weeks; the versions of each
model vary three settings
each. Equal tuning of the models with the rules covers the bent and kinked worlds, not the log-linear one. The Dominick's price
episodes are not used, because our explanation audit reports results on them. In robotics or driving the test needs a simulator or
logged interventions for the known answer; we have not run it there.

\section{Conclusion}\label{sec:conclusion}

A world model can be tested on actions it has never seen by holding them out by design, scoring them against a known
answer and giving its structure to simpler models. In retail and in a dosing world, forecast accuracy is a fair guide
between kinds of model and a poor guide between versions of one world model where actions follow need. Training on
actions does not help; the regular price and three price rules do. With them a simple model is best when its
curve can match the response; in our kinked world the world model with the rules led the simpler models we tested.
On real data the rules help on deep cuts, not lasting moves.

\section{Reproducibility and compute}\label{sec:repro}

\texttt{make} in the paper folder rebuilds every number, table and figure from the run records, identically from a
fresh copy of the repository; runs are \texttt{scripts/paper3\_*.py}. The \nFits{} fits cost about US\$135 on Modal
CPU. The plan and its ten amendments are in \texttt{docs/PAPER3\_PLAN.md}.

\section{Data and code availability}\label{sec:avail}

The generator, models, scripts (\texttt{scripts/paper3\_*.py}) and run records (\texttt{results/paper3/}) are at commit
\texttt{e8e021d} of \url{https://github.com/karthik-iyer-au/retail-world-models} (branch \texttt{paper3}); the manuscript
source is in \texttt{paper3/} on the same branch, and \texttt{make -C paper3} rebuilds it from the saved run records. The shared forecasting code is also archived with the companion study~\cite{rwm},
under the same licence.

\section*{Use of generative AI}
Claude (Anthropic) was used to write code and draft text under the author's direction. The author designed the study,
fixed the plan and its amendments, checked the results and takes responsibility for the paper.

\input{references}
\appendix

\section{Plan and amendments}\label{app:plan}

The plan was fixed on 16 November 2026 before any run: the worlds, the 25 versions per world, the held-out plans,
the measures and the test of \S\ref{sec:setup}. \emph{Amendment 1}, written after the first results and before its
runs, added the action-learning versions, the recorded regular price for every model and the bent world.
\emph{Amendment 2}, written before its runs, added the split model and test and learn; it was revised before any
test-and-learn result was read, to keep test cuts inside the merchants' range and to add the world model with the
price rules. \emph{Amendment 3}, written before its runs after a reviewer's challenge that the world model with the
rules might win only because its rules match the worlds, added the same rules to the log-linear model and to
LightGBM and the kinked world, and corrected the description of Equation~\eqref{eq:shape}. \emph{Amendment 4}, written before its runs after
the same reviewer noted that the controls used one version of each model and that versions are not independent
replications, added equal tuning for the three models with the rules, 28 more kinked worlds for the head-to-head,
and the statistics with the world as the unit. \emph{Amendment 5}, written before its runs at the author's request
for a real-data check, added the tests on Breakfast at the Frat and M5; \emph{amendment 6}, after its summary
results, added pure price cuts, clean lasting moves, the 20\% threshold and more origins; \emph{amendment 7} added The
Complete Journey; \emph{amendment 8}, at the author's request for a second domain, added the dosing world, and \emph{amendment 9}
the drug level built up from recent doses.
Each was written before the analyses it added, and every result is reported. The correlation within each model class and the cap on the error of new plans were added in the
analysis; the cap counts a wrong sign as a twentyfold miss so the measure does not move as versions are added.

\section{What was tried and dropped}\label{sec:dropped}

\paragraph{Test plans as deep as 60\%} The first version of test and learn drew test cuts up to 60\%. That puts the
held-out cuts inside the training range and tests memory, not transfer, so those runs were cancelled before any
result was read and the tests kept to the merchants' range (Appendix~\ref{app:plan}).

\paragraph{The world model given the regular price alone, and training for actions} Each left the error on new plans
no lower (\S\ref{sec:results}); both are worth revisiting with the price rules, or with histories that hold varied
actions, such as many retailers' data pooled.

\section{The dosing world}\label{app:dosing}

360 patients in 18 clinics over 260 days (\texttt{scripts/paper3\_dosing.py}). Severity follows an AR(1) process
(persistence 0.8) and multiplies the expected daily response by $e^{-0.5 s}$; the drug concentration is
$C_t = C_{t-1} e^{-\ln 2/3} + d_t$ (half-life 3 days) and multiplies it by $1 + C/(C + 0.5)$; counts are Poisson.
Doctors give a dose with probability $\sigma(-1.4 + k s)$, sized 10\% to 40\% of the maximum and larger when severity
is high, for dosing strength $k$. The dose reaches every model through the price columns (price $1 - 0.95d$, regular
price 1), so the models and rules of the retail study apply unchanged. Fit on days 1 to 252, forecast days 253 to 260.

The confounding mirrors the retail worlds. There, merchants plan cuts into weeks of high demand, so a model that
reads sales against price sees the planning as well as the response. Here, doctors dose patients who are sicker, so
a model that reads the response against the dose sees larger doses with lower responses and can learn that a higher
dose lowers the response; the dosing strength $k$ plays the part of planning strength. The five held-out plans
are a dose of 50\% or 60\% on the days doses were given (deeper than any given), a steady 20\% every day (a lasting
change), the same doses moved to each patient's days of lowest severity (doses off their usual pattern) and no dose.
Each is scored against the known expected response, as in retail. In amendment 9 the price column carries the drug
level built up from recent doses, $\min(C, 3)/3$, in place of the day's dose, which gives the models the state the
response depends on but not the severity behind the doctors' choices. The world is a test of the evaluation in a
second domain, not a model of any drug or disease.

\input{generated/tab_dosing}

\section{The price rules and their constraints}\label{app:shape}

In Equation~\eqref{eq:shape} the four parameters are a learned bias plus a linear map of the item and store
embeddings, each passed through a softplus ($b$ adds 0.02), so each is positive. The cut is $1 - p/r$ for planned
price $p$, clipped to $[0, 0.95]$; a week with no price gets no cut. The term in $\gamma$ enters the baseline as
well as the forecast under the plan, so a lasting regular-price move changes the baseline. The price and the regular
price are replaced by constants in every other input of the network. The option is \texttt{price\_shape} of the
state model and needs the regular price and the lift readout.

%% file: generated/tab_link.tex
\begin{table*}[t]
\centering
\caption{\textbf{Does better forecasting go with better simulation?} Rank correlation, within each world, between
forecast error on the merchants' plans and each measure (higher is worse for every measure, so a positive number
means better forecasters are also better at it), averaged over three seeds. Bold: the 95\% bootstrap interval excludes
zero. --: every version made the same choice, so there is nothing to rank. All versions: the test fixed in the plan (12 world model, 12 LightGBM and the log-linear model). Within a model
class: added after the first results.}
\label{tab:link}
\footnotesize\setlength{\tabcolsep}{4pt}
\begin{adjustbox}{max width=\textwidth}
\begin{tabular}{llrrrrrrrrr}
\toprule
 & & \multicolumn{3}{c}{All 25 versions} & \multicolumn{3}{c}{World model versions} & \multicolumn{3}{c}{LightGBM versions} \\
\cmidrule(lr){3-5}\cmidrule(lr){6-8}\cmidrule(lr){9-11}
World & Planning & New plans & Price choice & Stock & New plans & Price choice & Stock & New plans & Price choice & Stock \\
\midrule
Log-linear world & 0 & \textbf{0.51} & 0.15 & \textbf{0.32} & \textbf{0.64} & -- & \textbf{0.83} & 0.33 & \textbf{$-$0.44} & \textbf{0.88} \\
 & 0.5 & \textbf{0.33} & \textbf{0.44} & 0.14 & 0.18 & -- & \textbf{0.79} & 0.38 & $-$0.33 & \textbf{0.87} \\
 & 1 & \textbf{0.40} & 0.15 & 0.22 & $-$0.31 & -- & \textbf{0.67} & 0.23 & \textbf{$-$0.67} & \textbf{0.75} \\
 & 1.5 & \textbf{0.34} & 0.14 & \textbf{0.27} & \textbf{$-$0.35} & -- & \textbf{0.85} & 0.24 & \textbf{$-$0.74} & \textbf{0.87} \\
Bent world & 0 & \textbf{0.32} & 0.13 & \textbf{0.69} & \textbf{0.43} & \textbf{0.30} & \textbf{0.86} & 0.40 & \textbf{$-$0.36} & \textbf{0.84} \\
 & 0.5 & \textbf{0.54} & \textbf{0.55} & \textbf{0.84} & $-$0.25 & -- & \textbf{0.92} & 0.23 & $-$0.29 & \textbf{0.82} \\
 & 1 & \textbf{0.48} & \textbf{0.56} & \textbf{0.87} & \textbf{$-$0.69} & -- & \textbf{0.90} & 0.27 & \textbf{$-$0.44} & \textbf{0.89} \\
 & 1.5 & \textbf{0.66} & \textbf{0.46} & \textbf{0.89} & $-$0.05 & -- & \textbf{0.88} & \textbf{0.62} & \textbf{$-$0.52} & \textbf{0.91} \\
\bottomrule
\end{tabular}
\end{adjustbox}
\end{table*}

%% file: generated/tab_remedies.tex
\begin{table*}[t]
\centering
\caption{\textbf{Every model and remedy in the three worlds.} Medians over the 12 worlds of each kind (four planning
strengths, three seeds). Forecast: item-by-store RMSSE on the merchants' own plan in the held-out weeks. New plans:
mean over the eight held-out plans of $|\log(\text{predicted change}/\text{true change})|$. Price choice: revenue
given up by the regular-price move the model picks, \% of the best. Stock: summed absolute error of ordering to the
forecast, over expected sales. Rows marked + change one thing in the row above; all rows after the first of each
model, and the split rows, were added after the first results (Appendix~\ref{app:plan}). The kinked world was run
for the versions shown.}
\label{tab:remedies}
\footnotesize\setlength{\tabcolsep}{3.2pt}
\begin{adjustbox}{max width=\textwidth}
\begin{tabular}{lrrrrrrrrrrrr}
\toprule
 & \multicolumn{4}{c}{Log-linear world} & \multicolumn{4}{c}{Bent world} & \multicolumn{4}{c}{Kinked world} \\
\cmidrule(lr){2-5}\cmidrule(lr){6-9}\cmidrule(lr){10-13}
Model & Forecast & New plans & Price & Stock & Forecast & New plans & Price & Stock & Forecast & New plans & Price & Stock \\
\midrule
World model & 0.362 & 1.06 & 9.5 & 0.26 & 0.475 & 1.39 & 9.6 & 0.26 & 0.544 & 0.92 & 9.5 & 0.21 \\
\quad + JEPA pretraining & 0.364 & 1.06 & 9.5 & 0.26 & 0.471 & 1.38 & 9.6 & 0.26 &  &  &  &  \\
\quad + latent check & 0.364 & 1.06 & 9.5 & 0.26 & 0.469 & 1.35 & 9.6 & 0.25 &  &  &  &  \\
\quad + latent check 7 ahead & 0.364 & 1.05 & 9.5 & 0.26 & 0.471 & 1.34 & 9.6 & 0.25 &  &  &  &  \\
\quad + adaptation at the plan date & 0.364 & 1.06 & 9.5 & 0.26 & 0.472 & 1.39 & 9.6 & 0.26 &  &  &  &  \\
\quad + recorded regular price & 0.339 & 1.65 & 0.0 & 0.21 & 0.459 & 2.14 & 0.0 & 0.22 & 0.542 & 1.87 & 0.0 & 0.20 \\
\quad + price rules & 0.343 & 0.39 & 0.9 & 0.23 & 0.444 & 0.27 & 0.4 & 0.19 & 0.547 & 0.30 & 0.2 & 0.19 \\
Split: world model \quad + log-linear & 0.340 & 0.17 & 0.0 & 0.20 & 0.469 & 0.71 & 0.0 & 0.27 & 0.544 & 0.48 & 0.0 & 0.22 \\
LightGBM & 0.355 & 0.93 & 4.2 & 0.28 & 0.450 & 0.89 & 2.7 & 0.22 & 0.546 & 0.72 & 1.7 & 0.20 \\
\quad + recorded regular price & 0.353 & 0.89 & 3.9 & 0.27 & 0.450 & 0.77 & 2.7 & 0.22 & 0.549 & 0.67 & 1.6 & 0.20 \\
\quad + price rules & 0.355 & 0.82 & 4.4 & 0.27 & 0.450 & 0.59 & 2.6 & 0.22 & 0.551 & 0.49 & 1.6 & 0.20 \\
Split: LightGBM \quad + log-linear & 0.339 & 0.18 & 0.0 & 0.20 & 0.467 & 0.72 & 0.0 & 0.28 & 0.548 & 0.49 & 0.0 & 0.22 \\
Log-linear & 0.354 & 0.82 & 9.5 & 0.24 & 0.470 & 1.22 & 9.1 & 0.29 & 0.544 & 0.76 & 5.3 & 0.22 \\
\quad + recorded regular price & 0.337 & 0.17 & 0.0 & 0.20 & 0.465 & 0.73 & 0.0 & 0.28 & 0.542 & 0.49 & 0.0 & 0.21 \\
\quad + price rules & 0.336 & 0.13 & 0.0 & 0.19 & 0.438 & 0.17 & 0.0 & 0.19 & 0.542 & 0.40 & 0.0 & 0.20 \\
\bottomrule
\end{tabular}
\end{adjustbox}
\end{table*}

%% file: generated/tab_h2h.tex
\begin{table*}[t]
\centering
\caption{\textbf{The three models with the price rules, head to head, with the world as the unit} (added after the
first results). Each cell: worlds in which the first model has the lower error on new plans, of all worlds (two-sided
sign test), then the median paired difference in error (negative favours the first model) with a 95\% interval from
a bootstrap over worlds. Base: one version of each model. Median and best: each model's median and best version on
the held-out plans over 12 world model, 12 LightGBM and 6 log-linear versions. The kinked base comparison has 40
worlds (seeds 101 to 110); the tuned comparisons have 12.}
\label{tab:h2h}
\footnotesize\setlength{\tabcolsep}{3.5pt}
\begin{adjustbox}{max width=\textwidth}
\begin{tabular}{lllll}
\toprule
World & Comparison & Base version & Median version & Best version \\
\midrule
Log-linear & World model vs Log-linear & 0/12 ($<$0.001); +0.21 [+0.15, +0.32] &  &  \\
 & World model vs LightGBM & 12/12 ($<$0.001); $-$0.46 [$-$0.50, $-$0.39] &  &  \\
 & Log-linear vs LightGBM & 12/12 ($<$0.001); $-$0.67 [$-$0.76, $-$0.59] &  &  \\
Bent & World model vs Log-linear & 1/12 (0.006); +0.12 [+0.09, +0.13] & 0/12 ($<$0.001); +0.17 [+0.14, +0.19] & 1/12 (0.006); +0.08 [+0.04, +0.10] \\
 & World model vs LightGBM & 12/12 ($<$0.001); $-$0.30 [$-$0.33, $-$0.24] & 12/12 ($<$0.001); $-$0.36 [$-$0.43, $-$0.31] & 12/12 ($<$0.001); $-$0.35 [$-$0.38, $-$0.30] \\
 & Log-linear vs LightGBM & 12/12 ($<$0.001); $-$0.41 [$-$0.47, $-$0.38] & 12/12 ($<$0.001); $-$0.53 [$-$0.60, $-$0.48] & 12/12 ($<$0.001); $-$0.43 [$-$0.48, $-$0.38] \\
Kinked & World model vs Log-linear & 36/40 ($<$0.001); $-$0.12 [$-$0.18, $-$0.09] & 12/12 ($<$0.001); $-$0.12 [$-$0.18, $-$0.09] & 12/12 ($<$0.001); $-$0.16 [$-$0.21, $-$0.12] \\
 & World model vs LightGBM & 38/40 ($<$0.001); $-$0.21 [$-$0.25, $-$0.18] & 11/12 (0.006); $-$0.27 [$-$0.34, $-$0.16] & 11/12 (0.006); $-$0.22 [$-$0.27, $-$0.12] \\
 & Log-linear vs LightGBM & 27/40 (0.038); $-$0.09 [$-$0.18, $-$0.01] & 8/12 (0.388); $-$0.13 [$-$0.23, +0.08] & 7/12 (0.774); $-$0.04 [$-$0.13, +0.16] \\
\bottomrule
\end{tabular}
\end{adjustbox}
\end{table*}

%% file: generated/tab_real.tex
\begin{table*}[t]
\centering
\caption{\textbf{Real data: error on the rarest plans that were actually run} (added after the first results;
Appendix~\ref{app:plan}). Summed absolute error over summed sales, median over the forecast origins; lower is
better. Deep cuts: every cut deeper than the threshold (30\% or 20\%) was removed from training and the held-out
weeks with such a cut are scored; pure: with no circular or display, so price is the only lever that changed.
Lasting moves: on the Frat, a recorded regular price 10\% or more from its last value, steady for four weeks before
the origin and through every held-out week; on M5, a move of 5\% or more in a constructed regular price. These are
sales that happened, not plans never run.}
\label{tab:real}
\footnotesize\setlength{\tabcolsep}{4pt}
\begin{adjustbox}{max width=\textwidth}
\begin{tabular}{lrrrrrr}
\toprule
 & \multicolumn{4}{c}{Deep cuts never seen in training} & \multicolumn{2}{c}{Lasting regular-price moves} \\
\cmidrule(lr){2-5}\cmidrule(lr){6-7}
 & Frat $>$30\% & Frat $>$30\%, pure & Frat $>$20\%, pure & Complete Journey $>$30\%, pure & Frat & M5 \\
Item-weeks / origins & 8,341 / 8 & 6,288 / 8 & 12,793 / 8 & 324 / 5 & 1,115 / 6 & 1,932 / 4 \\
\midrule
World model & 0.393 & 0.627 & 0.488 & 0.446 & 0.356 & 0.551 \\
\quad + recorded regular price & 0.402 & 0.717 & 0.510 & 0.466 & 0.392 & 0.541 \\
\quad + price rules & 0.431 & 0.558 & 0.473 & 0.439 & 0.378 & 0.532 \\
LightGBM & 0.407 & 0.531 & 0.426 & 0.499 & 0.368 & 0.539 \\
\quad + recorded regular price & 0.410 & 0.520 & 0.421 & 0.481 & 0.358 & 0.544 \\
\quad + price rules & 0.421 & 0.566 & 0.425 & 0.493 & 0.359 & 0.529 \\
Log-linear & 0.427 & 1.064 & 0.590 & 0.460 & 0.412 & 0.519 \\
\quad + recorded regular price & 0.410 & 1.291 & 0.673 & 0.441 & 0.433 & 0.520 \\
\quad + price rules & 0.410 & 1.265 & 0.504 & 0.453 & 0.437 & 0.520 \\
\bottomrule
\end{tabular}
\end{adjustbox}
\end{table*}

%% file: generated/tab_dosing.tex
\begin{table}[t]
\centering
\caption{\textbf{The dosing world} (added; Appendix~\ref{app:plan}). Medians over three seeds at each dosing strength
(how closely doses follow severity). Rows marked drug level are given the level built up from recent doses in
place of the day's dose (amendment 9). Error on new doses as in the retail study (capped at 3.00, a twentyfold miss or a
wrong sign); predicted change over true change for every planned dose raised to 60\% of the maximum.}
\label{tab:dosing}
\footnotesize\setlength{\tabcolsep}{3pt}
\begin{tabular}{lrrrrrrrr}
\toprule
 & \multicolumn{4}{c}{Error on new doses} & \multicolumn{4}{c}{Change ratio, 60\% dose} \\
\cmidrule(lr){2-5}\cmidrule(lr){6-9}
Strength & 0 & 0.5 & 1 & 1.5 & 0 & 0.5 & 1 & 1.5 \\
\midrule
World model & 0.75 & 3.00 & 3.00 & 3.00 & 2.88 & $-$0.89 & $-$2.19 & $-$2.44 \\
\quad + dose rules & 1.18 & 2.76 & 2.99 & 2.96 & 0.39 & 0.08 & 0.05 & 0.06 \\
LightGBM & 1.74 & 2.41 & 2.28 & 2.20 & 0.13 & $-$2.32 & $-$2.96 & $-$3.60 \\
\quad + dose rules & 1.71 & 2.58 & 2.62 & 2.68 & 0.15 & $-$0.64 & $-$0.73 & $-$0.59 \\
Log-linear & 1.01 & 2.91 & 2.67 & 2.69 & 1.39 & $-$0.67 & $-$1.43 & $-$1.89 \\
\quad + dose rules & 1.31 & 3.00 & 2.90 & 2.96 & 0.34 & 0.00 & 0.00 & 0.00 \\
World model, drug level & 0.80 & 2.55 & 2.53 & 2.65 & 2.62 & $-$0.30 & $-$1.14 & $-$1.53 \\
\quad + dose rules & 0.73 & 1.66 & 2.55 & 2.54 & 1.27 & 0.32 & 0.12 & 0.12 \\
LightGBM, drug level & 0.63 & 2.64 & 2.50 & 2.52 & 0.97 & $-$2.22 & $-$2.76 & $-$2.94 \\
\quad + dose rules & 0.64 & 2.59 & 2.88 & 3.00 & 0.95 & $-$0.40 & $-$0.38 & $-$0.17 \\
Log-linear, drug level & 0.93 & 3.00 & 3.00 & 2.96 & 2.75 & $-$1.04 & $-$2.00 & $-$2.30 \\
\quad + dose rules & 0.65 & 3.00 & 3.00 & 3.00 & 1.03 & 0.00 & 0.00 & 0.00 \\
\bottomrule
\end{tabular}
\end{table}